# Automatic Labeled LiDAR Data Generation based on Precise Human Model

Wonjik Kim, Masayuki Tanaka, Masatoshi Okutomi, Yoko Sasaki

*Abstract*— Following improvements in deep neural networks, state-of-the-art netwy depends on the training data. An issue with collecting training data is labeling. Labeling by humans is necessary to obtain the ground truth label; however, labeling requires huge costs. Therefore, we propose an automatic labeled data generation pipeline, for which we can change any parameters or data generation environments. Our approach uses a human model named Dhaiba and a background of Miraikan and consequently generated realistic artificial data. We present 500k+ data generated by the proposed pipeline. This paper also describes the specification oforks have been proposed for human recog- nition using point clouds captured by LiDAR. However, the performance of these networks strongl the pipeline and data details with evaluations of various approaches.

## I. INTRODUCTION

Deep neural networks have considerably impacted computer vision, robotics applications, among other fields. For example, various types of high-performance deep neural networks for pixel-wise segmentation of RGB images have been proposed [1]–[3]. Following its success in RGB image segmentation, segmentation networks for depth map and point cloud data have also been vigorously researched [4]–[7]. Human segmentation is a very important task in many types of robotics applications. Recently, Light Detection and Ranging (LiDAR) has become a powerful tool for human segmentation and detection. Therefore, in this study, we focus on human segmentation with three-dimensional point cloud data collected by the LiDAR.

Deep neural networks commonly require a considerable amount of manually labeled training data to achieve high performance. Collecting a sufficient amount of labeled data may incur massive costs in terms of both time and money. In this study, we develop a data generation pipeline that reduces the manual labeling cost. The proposed data generation pipeline comprises three steps: 1) background data collection, 2) human model building, and 3) LiDAR data generation with human labels. The data generation pipeline can provide depth maps, point cloud data with xyz coordinates, and ground truth human labels. The generated data already includes the ground truth human labels; therefore, we do not need a manual labeling process. Additionally, the proposed data generation pipeline can produce point cloud data with any size of human and any type of LiDAR by changing several parameters.

W. Kim, M. Tanaka, and M. Okutomi are with the Department of Systems and Control Engineering, School of Engineering, Tokyo Institute of Technology, Meguto-ku, Tokyo 152-8550, Japan (e-mail: wkim@ok.sc.e.titech.ac.jp; mtanaka@sc.e.titech.ac.jp; mxo@sc.e.titech.ac.jp).

W. Kim, M. Tanaka, and Y. Sasaki are also with Artificial Intelligence Research Center, National Institute of Advanced Industrial Science and Technology, Koto-ku, Tokyo 135-0064, Japan (email: y-sasaki@aist.go.jp).

Our goal is to accelerate research of learning-based human segmentation with point cloud data. For that purpose, we have generated a large amount of labeled depth maps (more than 500K maps) with the proposed data generation pipeline. We have trained several segmentation networks with the generated training data. The trained networks have been evaluated with actual data collected by a real LiDAR sensor.

This paper is organized as follows. We quickly review related works in section II. In section III, the entire data generation procedure is explained in detail with a description of each step. Section IV-A contains a brief explanation of segmentation, and three different networks. The specific policy of the training and evaluation methods are described in section IV. A conclusion is provided in section V.

All generated data and labeled real data are presented in following url. Trained network weight and test sample code also included.

http://www.ok.sc.e.titech.ac.jp/res/LHD/

## II. RELATED WORK

After the release of the Microsoft Kinect in 2010, several RGB-D datasets were published. RGB-D datasets for human recognition were also provided, such as for the re-identification of a person with RGB-D sensors [8], BIWI RGBD-ID dataset [9], and UPCV Gait dataset [10]. As the Kinect cannot measure depths greater than 10 [m], LiDAR sensors were employed to handle depths over 10 [m]. In addition, LiDAR sensors are used in auto driving technology. In this field, the KITTI dataset [11] is widely used by many researchers [12], [13]. However, KITTI only provided 93K+ depth data without labeling. Collecting labeled depth maps is still challenging. Pixel or point-wise labeling for 3D depth data is usually a challenging task that involves massive cost. Under the circumstances, the video game Grand Theft Auto was deployed to collect data [4], [14], [15]. This approach may reduce the cost of data construction, but there are still limitations. Grand Theft Auto is not designed for research purposes; therefore, we cannot control specific properties of the circumstances of the simulation, such as human body type and model deployment location.

To deal with these problems, we constructed a data generation pipeline. This allows us to change any parameters and environments when generating depth maps. Furthermore, the proposed generator makes human labels in the process, thereby incorporating the human task into the computation cost. As a result, we can generate data continuously if there are enough computational resources.

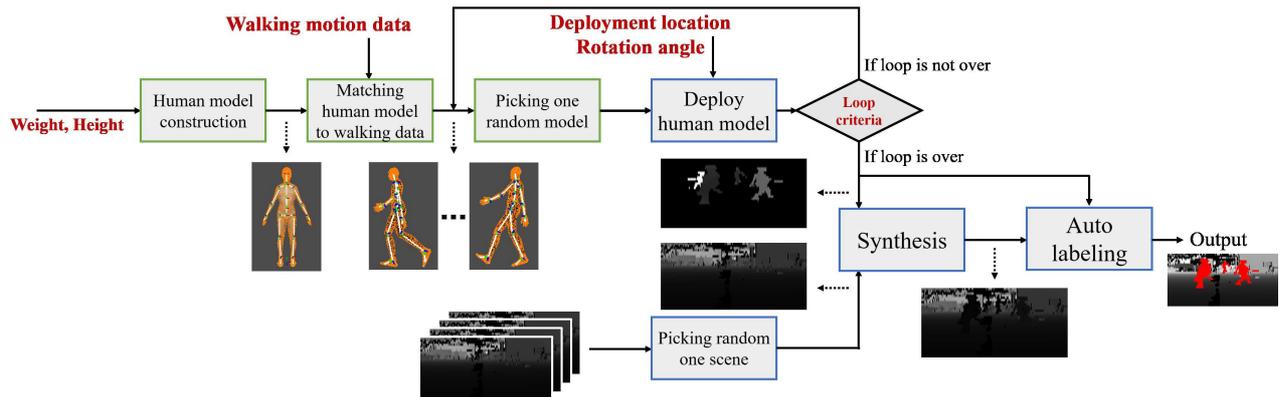

Fig. 1: Overview of the pipeline. Red words represent controllable parameters. (Green boxes: Human model building) Human models are built by weight and height, then they are combined with a human walking motion. (Blue boxes: LiDAR data generation) Random models are picked and deployed into the simulation field. Then, the depth map with human models is synthesized with a randomly picked background depth map. The depth map is labeled based on the information from the human model deployed location.

## III. PROPOSED DATA GENERATION PIPELINE

### A. Overview

One of the main contributions of this paper is the automatic generation of labeled depth map without involving manual label. The labeled depth map generation comprises three steps: 1) background data collection, 2) human model building, and 3) depth map synthesis with automatic labeling. The background depth maps without human were collected in advance. We collected 2,363 background depth maps at the 3rd floor of Miraikan, a science museum in Tokyo [16], using a velodyne HDL-32E LiDAR sensor [17]. Precise and realistic 3D human models were synthesized based on a digital human model called Dhaiba. The depth maps were generated by composing the collected background depth map and the 3D human model. Label data can be obtained because we know where the 3D human models are. This means that we can generate labeled depth maps automatically without manual input. This is one of advantages of the proposed approach. The details are described in the following subsections, and an overview of the pipeline is illustrated in Fig. 1.

### B. Human model building

One of the key components of the proposed data generation pipeline is the precise 3D human model. We employed a digital human model, called "Dhaiba", of DhibaWorks [18] to build precise 3D human models. The Dhaiba is a human body function model, and DhaibaWorks is a platform for producing digital human models based on Dhaiba. DhaibaWorks supports editing and visualizing basic models such as 3D meshes and skeletal structures, including human models with motion [19]. Using DhaibaWorks, we can easily generate a specific human model by setting the human parameters such as height, weight, and action status [20].

We assumed that the human model is walking during the depth map generation. Walking motion data is required to build an artificial walking human model. One period of walking data was collected using a motion capture system [21]. The walking motion data has 230 frames for a single walking motion. Then, it is easy to build artificial walking human mesh models for depth map generation. Details including motion capture system are described in Kobayashi's work [21].

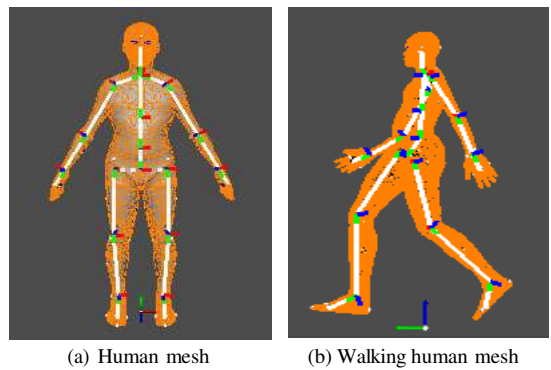

(a) Human mesh     (b) Walking human mesh

Fig. 2: Example of built human models based on Dhaiba

TABLE I: Human model specification

| Height (mm) | 1200 | 1400 | 1600 | 1700 | 1800 |
|---|---|---|---|---|---|
| Weight (kg) | 15 20 30 | 20 30 40 | 40 50 70 | 50 60 80 | 50 70 90 |

In this study, we take fifteen typical combinations of height and weight as summarized in Table I. We believe that these combinations cover a variety of scenarios. Now, we have 230 frames of walking motion data and fifteen combinations of height and weight. As such, a total of 3,450 different types of walking models can be generated. Figure 2 shows an example of built human models.

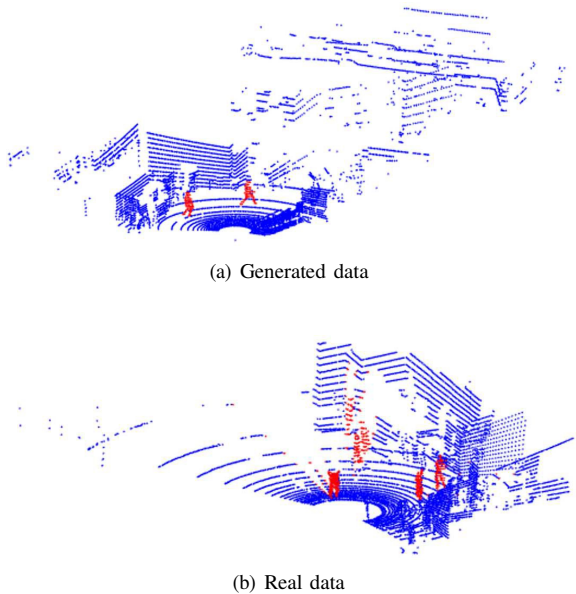

Fig. 3: Example of data. Blue points denote background and red points denote human, and real data labeled manually.

*C. LiDAR data generation with human labels*

In the LiDAR data generation step, the depth map of the built human model is first synthesized. Thereafter, the synthesized depth map of the human model and the randomly picked background depth map are combined to generate the training depth map for human segmentation.

To synthesize the human model depth map, we virtually position the LiDAR sensor at $(0, 0, z_s)$, where $z_s$ is the same height as the height of the sensor for background collection. We assumed that the LiDAR sensor was installed horizontally. Then, the human model depth map was synthesized for the given human model, virtually inserting the human model at $(x_s, y_s, 0)$, where $(x_s, y_s)$ is a randomly sampled position. The human model direction was randomly sampled. Once the geometrical positions of the LiDAR sensor and the human model were provided, the associated depth map can easily be synthesized.

After synthesizing the human model depth map, it is simply combined with the background depth map by pixel-wise minimum depth selection. In this step, human model farther than background is deleted by minimum depth selection. The depth map taken by the LiDAR sensor usually includes holes or missing pixels whose depth could not be measured. We leave these holes as they are for the synthesis process because these types of holes are equally obtainable in the real sensing process. In addition, the human labeling task can be simultaneously performed because we know which pixels correspond to the human model depth map.

The parameters in the LiDAR data generation are summarized in Table II. Figure 3 shows an example of the generated and real point cloud data. The point cloud data is converted from the depth map.

TABLE II: Data generation properties

| Description | Specification |
|---|---|
| Number of detector pairs | 32 |
| Limitation in horizontal scanning | 1024 |
| Vertical scanning range | +10.67 to -30.67° |
| Angular resolution in vertical | 1.33° |
| Angular resolution in horizontal | 0.2° |
| Human number in one image | 0 to 10 |
| Model located area | 0 to 25000[mm] from origin |
| Rotation angle of Dhaiba | 0 to 360 degrees |
| Output depth image size | 32 × 1024 × 1 |
| Output xyz map size | 32 × 1024 × 3 |
| Output label data size | 32 × 1024 × 1 |
| LiDAR position | (0, 0, 800) [mm] |

*D. Provision of generated data*

We generated 500K+ depth data generated by proposed pipeline. They contain depth, xyz coordinates, and human label in HDF5 format. We also provide specific information for each of 500K+ depth data in the shape of an xml file. Xml files contain a human number in the depth scene, location, weight, and height of each Dhaiba. The proposed pipeline allows us to control various properties easily. We also generated the 10K dataset described in section IV-C.2, 1K dataset for each property of the Dhiba described in section IV-C.3, 1K dataset for different backgrounds described in section IV-C.4, and 0.1K manually labeled real data.

## IV. EVALUATION

We used 47K generated data for training and 0.1K real data with manual labeling for validation and estimation. A total of 24101 human models were deployed in the 47K generated data. The model ratio with respect to distance was 74.45% between 0 to 5m, 14.78% between 5 to 10m, 4.726% between 10 to 15m, 3.301% between 15 to 20m, and 2.734% between 20 to 25m. Far models can easily be occluded if one human model is placed in close proximity to the LiDAR. This could be one reason the model ratio with respect to distance between 0 and 5 [m] is up to 74.45%.

In this section, four different networks are described in section IV-A, and the training method shown in IV-B. Section IV-C contains IV-C.1: the effect of train data number, IV-C.2: the effectiveness of precision of the human model, IV-C.3: a validation of different combinations of height and weight, IV-C.4 . Then, the result of four networks for generated and real data is shown in IV-D.

*A. Segmentation neural networks*

Semantic segmentation has been researched with various approaches for applications. The wide range of its application includes scene understanding, depth analysis, and autonomous driving. Semantic pixel-wise labeling has been gaining considerable interest due to improvements in deep learning [2], [22]–[24]. Four different networks were selected in this paper to evaluate the effectiveness of segmentation networks in the generated human depth maps.

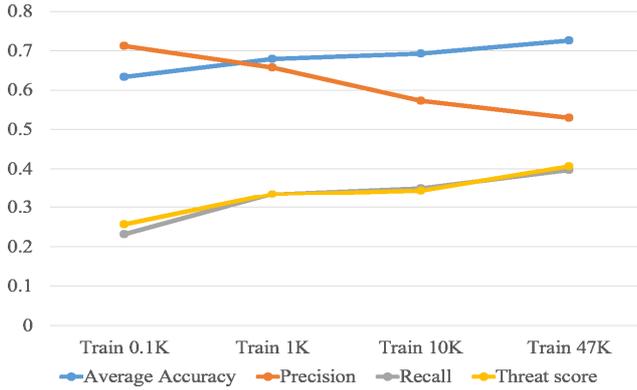

Fig. 4: Effect of training data number

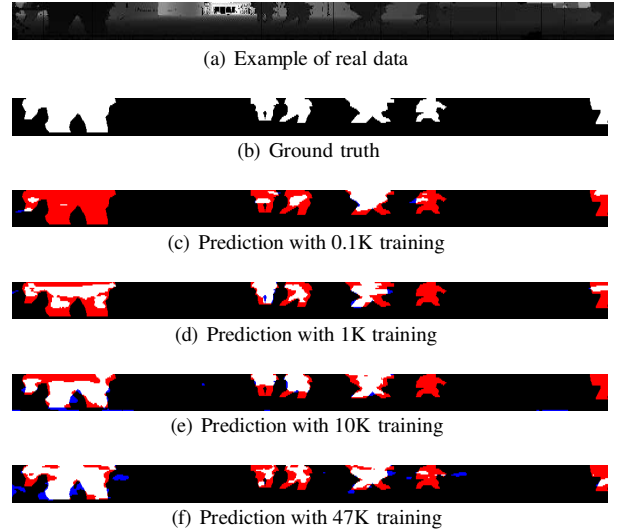

(a) Example of real data
(b) Ground truth
(c) Prediction with 0.1K training
(d) Prediction with 1K training
(e) Prediction with 10K training
(f) Prediction with 47K training

Fig. 5: Example of prediction result. In (b), white area denotes human label. From (c) to (f), white area denotes true positive, red area denotes false negative, while blue area denotes false positive.

*1) Fully convolutional network:* The Fully Convolutional Network (FCN) for segmentation was proposed by Long [1]. Classification networks usually use fully connected layers in the end of the network to compute class scores. On the contrary, FCN changed the fully connected layer to a convolutional layer, and the output of the network also changed from a predicted score to a heatmap. Thus, the classification network can easily be changed to a segmentation network using the concept of FCN.

*2) U-net:* U-net was proposed by Ronneberger [23].The U-net architecture includes path contracting for capturing context and a symmetric expanding path. These paths allow the network to enable precise localization. This network won the ISBI cell tracking challenge in 2015 by a large margin, and the authors suggested that U-net could be trained end-to-end from limited data with high performance.

*3) Fully convolutional DenseNets:* Fully convolutional DenseNets were proposed by Jégou [25]. They were inspired by Densely Connected Convolutional Networks (DenseNets) [26]. DenseNets show good performance in image classification, and the concept of DenseNets is based on the observation that if each layer is directly connected to every other layer in a feedforward manner, then the network will be more accurate and easier to train. Jégou proposed an extension to the concept of DenseNets to handle the problem of semantic segmentation. In this paper, we refer to this network as FCDN.

*4) Pointnet:* Pointnet was proposed by Charles [5]. Pointnet is designed for handling unordered point sets in 3D and is not affected by rotation or translation. In addition, by adapting a transform network into Pointnet, it accomplished gaining interactions among points and capturing local structures from neighbor points. Pointnet has two versions; classification and segmentation. The architectures of both versions show differences in the latter half. In this paper, we employed Pointnet as a segmentation network.

*B. Training policy*

We trained all networks with 500 epochs. In each epoch, we randomly picked 500 scenes from 47K datasets. We set a different batch size, chosen experimentally, for each network; 25 for FCN, 10 for U-net, 16 for PointNet, and 5 for FCDN. We then coordinated the step for epoch as $batchsize \times stepforepoch = 500$, except for PointNet. As PointNet takes a one-dimensional vector, we divided one scene into 16 vectors. Therefore, PointNet also trained with 500 scenes in one epoch, with a batch size of 16. We also employed Adam [27] with learning rate = 0.001 and decay = 0.001 for the optimizer and categorical cross-entropy for the objective function.

From the Table II, the size of the input image is $32 \times 1024$. Therefore, we revised the Fully Convolutional Network, U-net, and FCDN algorithms in pooling layers. As the height of the input image is only 32, we only pooled input data in the width direction. For Pointnet, input images were divided vertically into 16 pieces of images and dealt with 2048 pixels as points for input data. For FCDN, we also revised the pooling layer to a width-pooling layer, the number of layers per dense block to 3, and the growth rate to 8.

*C. Generated LiDAR data evaluation*

We analyzed the effectiveness of the produced data and verified them using various approaches. In this section, Avg. Acc denotes average of background accuracy and human accuracy. In addition, Avg. Acc, Precision, Recall, and Threat score [28] were calculated as positive as human. In addition, we used that the real data described in table III to compare and evaluate performances.

*1) Effect of training data number:* In general, the performance of the learning-based approach can be improved by increasing the volume of the training data, especially in the field of deep neural networks. However, in practice, it is difficult to collect a huge volume of training data due to the labeling cost. One of the advantages of the proposed depth map generation pipeline is automatic labeling. In this sense, the labeling cost of the proposed pipeline is much lower than that of manual labeling. Here, we clarify the effect of the number of training examples. We prepared four different sizes of training dataset, 0.1K, 1K, 10K, and 47K. Then, the FCN networks were trained with the four datasets. The trained networks were evaluated with the labeled real data.

TABLE III: Real data for evaluation

| Description | Specification |
|---|---|
| Location | Miraikan [16] |
| Date | 2017.Aug.04 |
| LiDAR model | Velodyne HDL32E |
| LiDAR height from ground | 800 [mm] |
| Data number | 100 |
| Annotation method | Manual |

TABLE IV: Comparison of results from different training data

| | Walking model | Standing model |
|---|---|---|
| Avg. Acc | 0.6922 | 0.6058 |
| Precision | 0.5729 | 0.7271 |
| Recall | 0.3504 | 0.2047 |
| Threat score | 0.3448 | 0.2061 |

TABLE V: Comparison of results from different body types

| | 1000mm, 12kg | 1500mm, 50kg | 2000mm, 80kg |
|---|---|---|---|
| Avg. Acc | 0.7747 | 0.7433 | 0.7423 |
| Precision | 0.4409 | 0.5403 | 0.5437 |
| Recall | 0.3886 | 0.3896 | 0.3814 |
| Threat score | 0.3635 | 0.4152 | 0.3879 |
| | 1400mm, 20kg | 1600mm, 50kg | 1700mm, 80kg |
| Avg. Acc | 0.7507 | 0.7771 | 0.7764 |
| Precision | 0.5110 | 0.5119 | 0.5716 |
| Recall | 0.3935 | 0.4131 | 0.4285 |
| Threat score | 0.4097 | 0.4142 | 0.4416 |

TABLE VI: Comparison of results from different backgrounds

| | Different background | Same background |
|---|---|---|
| Avg. Acc | 0.6556 | 0.7095 |
| Precision | 0.1297 | 0.7047 |
| Recall | 0.3515 | 0.4285 |
| Threat score | 0.1018 | 0.3919 |

Figure 4 shows the evaluation results. We can see that the average accuracy, recall, and threat score were improved by increasing the volume of the training data. However, precision shows a different tendency. In case of training with 0.1K data, the network could only be regarded as human when it has high confidence. Therefore, the average accuracy, recall, and threat score of the training 0.1K data were lower than those of any other cases but the precision was greater than 0.7. The network trained with 47K data shows higher performance than other cases in average accuracy, recall, and threat score, except for precision.

Figure 5 shows examples of the prediction for real data. From figure 5-(c) to figure 5-(f), it can be seen that increasing the number of the training data elevates the true positive number. Even the false positive rises as well, undetection of humans (false negative) is much more crucial than misdetection (false positive) in auto driving. Then, the increase in the recall with average accuracy and threat score can be considered entire improvements. In addition, the threat score denotes the intersection over union (IoU) in this experiment because there were only two labels. As a result, the increasing tendency of the threat score with respect to the number of training data represents improvement in the IoU. Following these reasons, we can conclude that a large volume of data helps to improve performance.

*2) Effectiveness of precision of human model:* We have proposed depth map generation based on a precise 3D human model. Here, we experimentally validate the effectiveness of the precise 3D human model. In the proposed pipeline, we assume that humans heights and weights are 15 different combinations and that the human is walking. Then, we build the precise 3D human model. For comparison, we built a human model assuming that the human height and weight was only 1600 mm-60 kg and that the human was standing toward the LiDAR, as shown in Fig. 2-(a). We call this human model the standing model. We generated 10K datasets with the human walking model of the proposed generation pipeline and with the standing model for comparison. Then, the FCN networks were trained with two training datasets. The two trained networks were evaluated using the manually labeled real data. Table IV shows the evaluation results. The comparison experimentally demonstrates that the precise human model improves the performance of the network.

*3) Validation of different combinations of height and weight:* We assumed 15 combinations of human heights and weights, as summarized in Table I. However, real human heights and weights were not a discrete 15 combinations. In this section, we test the FCN network with validation data with combinations of height and weight not used in the training. We picked combinations of 1000 mm-12 kg, 1500 mm-50 kg, and 2000 mm-80 kg for validation. The metrics used for the validation data are summarized in Table V, where we used the FCN network for validation. We also picked combinations of 1400 mm-20 kg, 1600 mm-50 kg, and 1700 mm-80 kg for comparison from Table I. Note that the training data does not include the validation data. From Table V, the performances for 1000 mm-12 kg, 1500 mm-50 kg, and 2000 mm-80 kg are comparable to those from Table V. Then, we can say that the trained networks sufficiently generalized in terms of the combinations of human heights and weights.

*4) Background depth map:* The performance of the human segmentation depends on the background scene. In this paper, we mainly focus on a specific site, namely Miraikan.

TABLE VII: Result of the networks for human segmentation

| Network | | FCN [1] | | | U-net [23] | | | FCDN [25] | | | Pointnet [5] |
|---|---|---|---|---|---|---|---|---|---|---|---|
| Input | | depth | xyz | depth+xyz | depth | xyz | depth+xyz | depth | xyz | depth+xyz | xyz |
| Generated data | Avg. Acc | 0.7529 | 0.7199 | 0.7160 | 0.5898 | 0.5696 | 0.5847 | 0.4990 | 0.5090 | 0.5870 | 0.4978 |
| | Precison | 0.5292 | 0.4776 | 0.5112 | 0.5441 | 0.3481 | 0.4213 | 0.0328 | 0.0564 | 0.0577 | 0.0443 |
| | Recall | 0.3973 | 0.3388 | 0.3107 | 0.1660 | 0.1477 | 0.1840 | 0.0064 | 0.0086 | 0.7707 | 0.1392 |
| | Threat score | 0.4062 | 0.3401 | 0.3565 | 0.1584 | 0.1268 | 0.1558 | 0.0035 | 0.0282 | 0.0568 | 0.0295 |
| Real data | Avg. Acc | 0.7095 | 0.5661 | 0.6178 | 0.5396 | 0.6364 | 0.6070 | 0.4990 | 0.5088 | 0.6050 | 0.5035 |
| | Precision | 0.7047 | 0.6623 | 0.4835 | 0.5831 | 0.2620 | 0.6126 | 0.0723 | 0.1369 | 0.1239 | 0.0719 |
| | Recall | 0.3508 | 0.1345 | 0.2238 | 0.0785 | 0.2948 | 0.1894 | 0.0031 | 0.0349 | 0.7848 | 0.0865 |
| | Threat score | 0.3919 | 0.1310 | 0.2146 | 0.0802 | 0.2003 | 0.2063 | 0.0035 | 0.0278 | 0.1194 | 0.0487 |

We collected the background data in advance. Here, we experimentally evaluated the effect of the background in the training. For that purpose, we collected outdoor background scenes, while the target scene is the inside of the building. Then, we generated the training data with the proposed pipeline for the outdoor background scene. After we trained the FCN networks with the training data for outdoor background scenes and Miraikan background scenes, the trained networks were evaluated with real depth maps taken at Miraikan. The evaluation results are summarized in Table VI. If the training background scenes were different from the target background, the performance of the network would be degraded, as expected. Therefore, our future works will include the collecting of various types of background data.

*D. Benchmarking existing networks*

We evaluated four different types of networks learning with the training data generated by the proposed pipeline. For the FCN, U-net, and FCDN networks, we prepared three different types of input: the depth data only, the xyz data only, and the depth-and-xyz data. We used the xyz data for the input of the Pointnet due to the network structure of the Pointnet. The manually labeled 0.1K real data and the 1K test data generated by the proposed pipeline were used for the evaluation. Note that the training data does not include the test data.

Table VII shows the evaluation results. From these comparisons, the FCN network with the depth information generally resulted in high performance, except for recall in the real data, precision, and recall in the generated data. In terms of recall, the FCDN network with depth-and-xyz information was the best. It could be observed that if a network shows any metric over 0.5 except precision in the test data generated by the proposed pipeline, that network also shows a value over 0.5 for that metric with the real data. Similarly, if the metric except precision for the generated test data is less than 0.5, the metric for the real data is also less than 0.5. We can say that evaluation with the generated test data helps predict the effectiveness for the real data. An example of prediction with FCN trained by depth and xyz coordinates is illustrated in Figure 6.

V. CONCLUSION

In this paper, we proposed a fully automated data generation pipeline for human detection using LiDAR. With

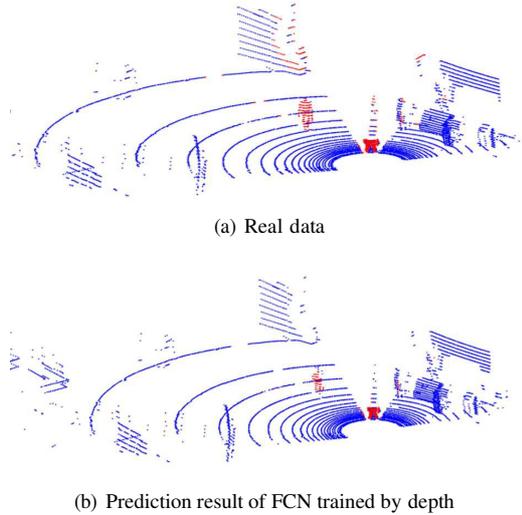

(a) Real data

(b) Prediction result of FCN trained by depth

fig. 6: Example of result. Human label is colored red while background label is colored blue.

this process, we can easily generate labeled data with any properties for LiDAR. Several approaches were taken to evaluate the generated data and compare the approaches. Following the result, we concluded that our generated data improve training for human detection. We presented 0.1K labeled real datasets and 500K+ generated datasets with human labels. We hope this dataset will support studies in many fields of robotics and computer vision.

We have considered three points for improving our work. The first point is in relation to the human model; although we used a confirmed method to produce the human model, it is not entirely representative of the real world. Then, we will try to consider clothes, backpacks, and other conditions for accurate simulations. The second point regards the background. Only data from Miraikan was used in the current investigation; in the future, we will employ other backgrounds for diversity of data generation. The third point is with regard to a balanced human ratio. Owing to occlusion, the current model ratio is unbalanced with respect to the distance. Filtering by distance or collecting data to balance the ratio may solve this problem.